\documentclass{iserc}

\usepackage[vlined,linesnumbered,ruled]{algorithm2e}
\usepackage{graphicx}
\usepackage{caption}
\usepackage{subcaption}
\usepackage{multirow}

\conference{Proceedings of the 2017 Industrial and Systems Engineering Conference\\
K. Coperich, E. Cudney, H. Nembhard, eds.}
\title{Collaborative Filtering using Denoising Auto-Encoders for Market Basket Data}
\author{
Andres G. Abad and Luis I. Reyes-Castro\\Escuela Superior Politecnica del Litoral (ESPOL)\\Guayaquil-Ecuador\\
\vspace{0.3cm}
}
\authorlist{Abad and Reyes-Castro}
\abstractID{3251}

\begin{document}

\maketitle

\begin{abstract}
Recommender systems (RS) help users navigate large sets of items in the search for ``interesting'' ones. One approach to RS is Collaborative Filtering (CF), which is based on the idea that \textit{similar} users are interested in \textit{similar} items. Most model-based approaches to CF seek to train a machine-learning/data-mining model based on sparse data; the model is then used to provide recommendations. While most of the proposed approaches are effective for small-size situations, the combinatorial nature of the problem makes it impractical for medium-to-large instances. In this work we present a novel approach to CF that works by training a Denoising Auto-Encoder (DAE) on corrupted baskets, i.e., baskets from which one or more items have been removed. The DAE is then forced to learn to reconstruct the original basket given its corrupted input. Due to recent advancements in optimization and other technologies for training neural-network models (such as DAE), the proposed method results in a scalable and practical approach to CF. The contribution of this work is twofold: (1) to identify missing items in observed baskets and, thus, directly providing a CF model; and, (2) to construct a generative model of baskets which may be used, for instance, in simulation analysis or as part of a more complex analytical method.
\end{abstract}

\section*{Keywords}
Recommender systems, collaborative filtering, denoising auto-encoders, market basket data, retail firms
\section{Introduction}


Recommender systems (RS) help users navigate large sets of items in the search for ``interesting'' ones.  In retail firms, RS work by providing recommendations based on the analysis of sparse transactional data: market basket data. On-line retailers were one of the first sectors to adopt RS; its application was popularized by Amazon's ``\textit{Customers who bought this item also bought}'' feature \cite{linden_amazon.com_2003}. In bricks-and-mortar stores, RS have been successfully applied to, for instance, designing up-selling strategies through customized discount coupons and targeted marketing campaigns. One approach to RS is Collaborative Filtering (CF), which is based on the idea that \textit{similar} users are interested in \textit{similar} items. 

In the model-based approach to CF, data is used to train a machine-learning/data-mining model. Supervised-learning models proposed for CF include regression models, as in \cite{vucetic_collaborative_2005} where a regression model was proposed for predicting user's ratings; and classification models, as in \cite{basu_recommendation_1998} where an inductive approach for classification was applied. In \cite{lee_classification-based_2005}, a logistic regression model together with a PCA for dimensionality reduction were used. Other supervised-learning approaches proposed for CF include bayesian classifiers \cite{miyahara_collaborative_2000} and belief networks \cite{su_collaborative_2006}.


Unsupervised-learning models proposed for CF include clustering techniques, such as in \cite{gong_efficient_2010} where $k$-means clustering was used. In \cite{shani_mdp-based_2005} the CF problem was addressed as a sequence of decisions, where the optimal policy was learned using a Markov Decision Process (MDP); in \cite{hofmann_latent_2004} Latent Semantic Analysis was used for CF, reporting higher accuracy and constant time prediction as two of the main advantages of their method. Another unsupervised-learning proposed approach to CF is to estimate the frequency of occurrence of combinations of items in the search of interesting \textit{association rules} \cite{agrawal_mining_1993}. A direct application of this type of analysis is the quantification of the complementary and supplementary relationship between items and their use for CF.

From a generative-model perspective, one may wish to learn the binary multivariate distribution of variables (representing the presence of items in each basket) to arrive to recommendations by performing probabilistic inference. In \cite{heckerman_dependency_2000} a graphical model, termed the dependency network, was used to estimate the conditional probability of each item given the others; probabilistic inference was made over the unconditional joint distribution using a Gibbs-sampling mechanism. In \cite{hruschka_analyzing_2014} a Restricted Boltzmann Machine (RBM) was used to learn the binary distribution of items, from which conditional-independence assumptions allowed for efficient estimation of the cross-category effect. In general, however, learning probability distributions with graphical models is intractable because of the normalization requirement.


While all these approaches are effective for small-size situations, the combinatorial nature of the problem makes it impractical for medium-to-large instances. In this work we present a novel approach to CF that works by training a Denoising Auto-Encoder (DAE) model on corrupted baskets, i.e., baskets from which one or more items have been removed. The DAE is then forced to learn to reconstruct the original basket given its corrupted input. The contribution of this work is twofold: (1) to identify missing items in observed baskets and, thus, directly providing a CF model; and, (2) to construct a generative model of baskets which may be used, for instance, in simulation analysis or as part of a more complex analytical method.

Due to recent advancements in optimization and other technologies for training neural-network models, the proposed method results in a scalable and practical approach to CF. We train the model using the Adam algorithm, a recently proposed optimization method for effectively training neural networks. Model implementation and training was performed using TensorFlow library, version 0.11.0, on a linux machine endowed with an NVIDIA GK520 GPU with 3,072 cuda cores.

For illustrating the proposed methodology, we used a publicly-available data set $\mathcal{S}$ consisting of 9,835 baskets on $p=10$ item categories \cite{hahsler_implications_2006}. We represent each basket by a binary vector $\mathbf{x}=[x_1,\dots,x_p]$, with $x_i$ equal to 1 if item $i$ belongs in the basket, and 0 otherwise. A corrupted basket is represented by $\mathbf{\tilde{x}}=[\tilde{x}_1,\dots,\tilde{x}_p]$, with similar definition for the $\tilde{x}_i$'s. For training and evaluating the model we partition data set $\mathcal{S}$ into training set $\mathcal{T}$ (consisting of  6,885 baskets) and evaluation set $\mathcal{E}$ (consisting of 2,950 baskets.)

\section{Denoising Auto-Encoders for Basket Modeling}\label{model}


DAE are neural-network models that are trained to denoise a corrupted signal. In this section we describe its application to the analysis of market basket data in the context of CF.
\subsection{Denoise Auto-Encoder Structure}

\begin{figure}[ht]
        \centering
        \begin{subfigure}[h]{.3\textwidth}
                \includegraphics[width=1.15\textwidth]{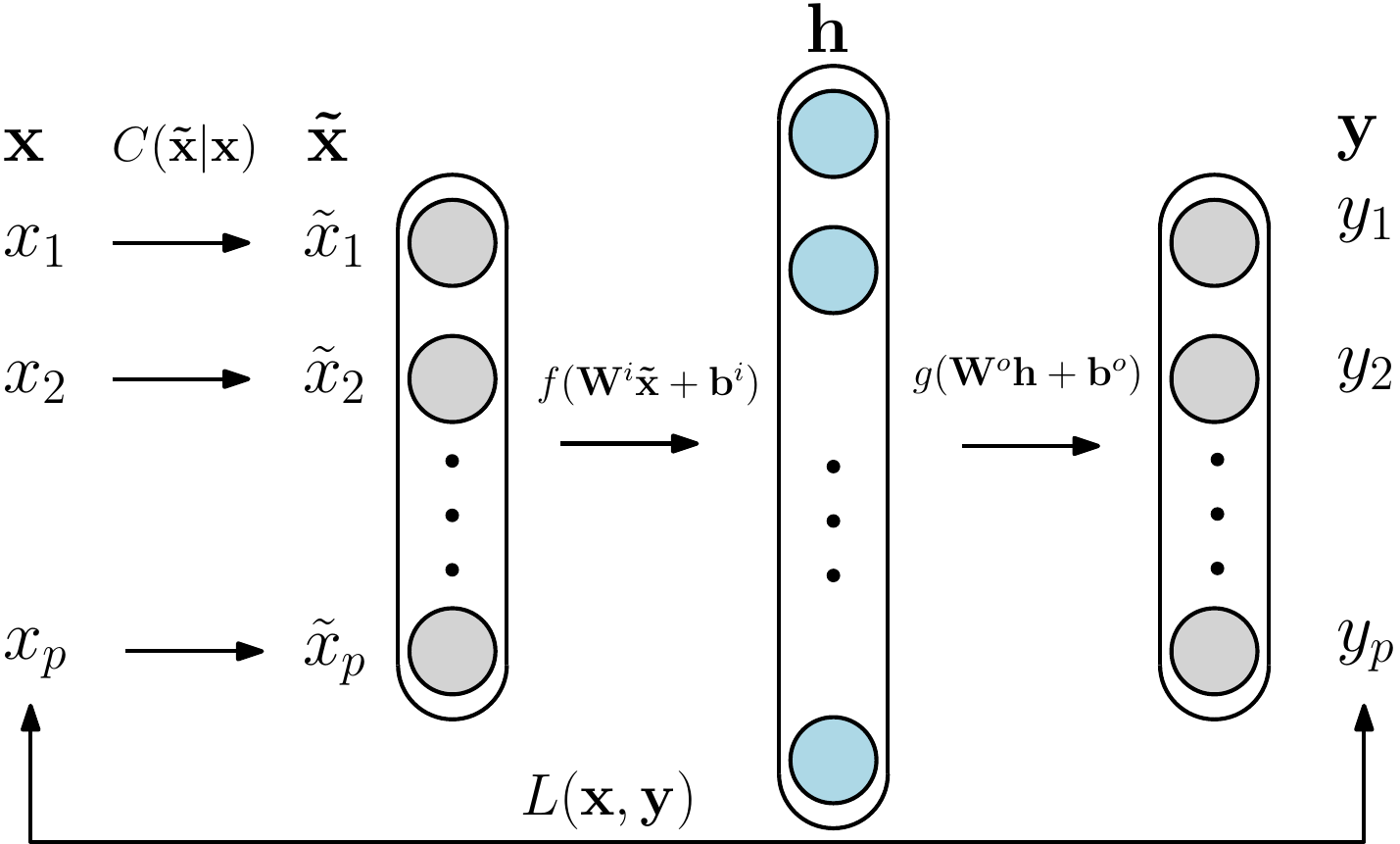}
                \caption{}
                \label{DAE}
        \end{subfigure}
        ~~~ 
        \begin{subfigure}[h]{.3\textwidth}
                \includegraphics[width=\textwidth]{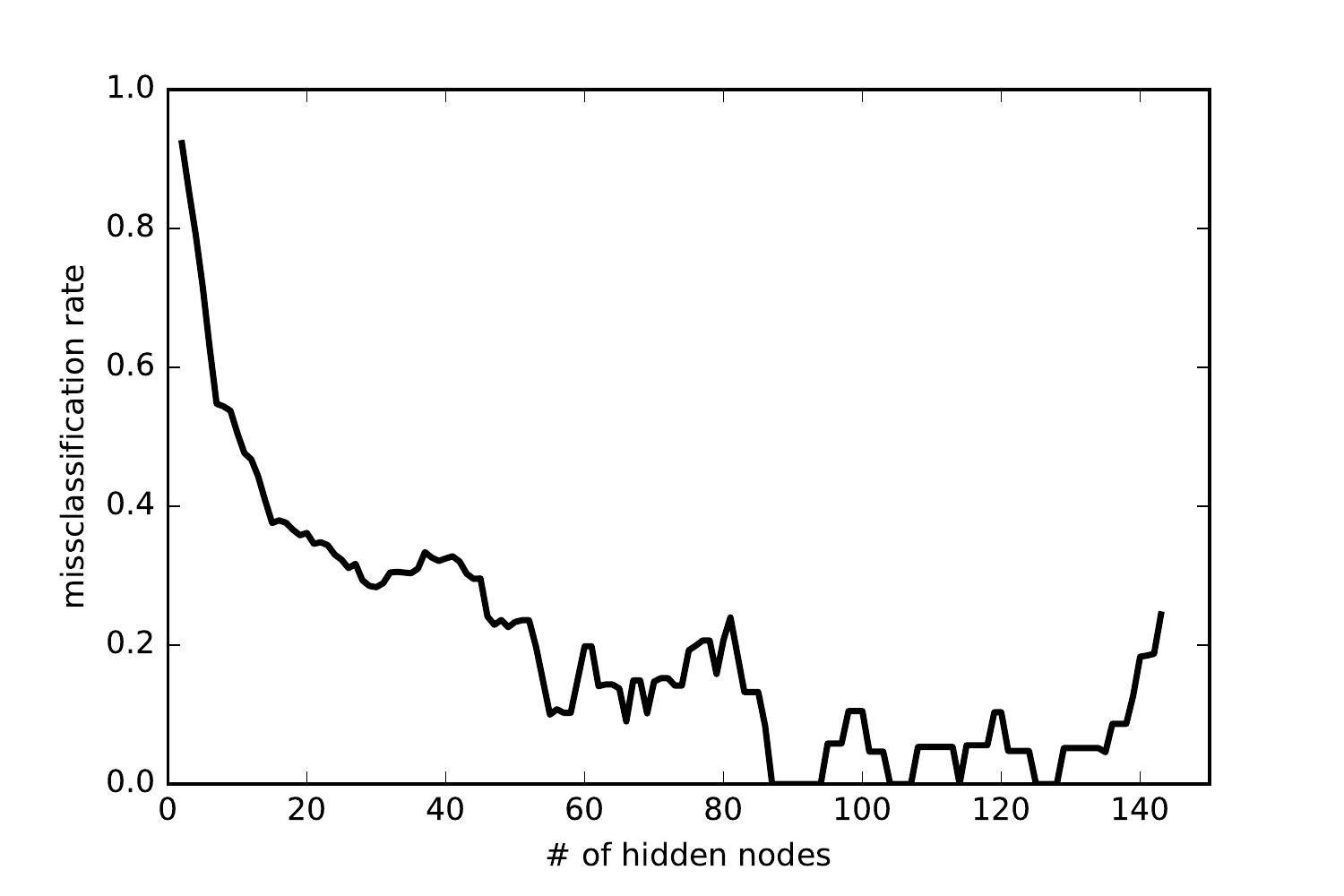}
                \caption{}
                \label{fig:missclassification}
        \end{subfigure}
        ~ 
        \begin{subfigure}[h]{.3\textwidth}
                \includegraphics[width=\textwidth]{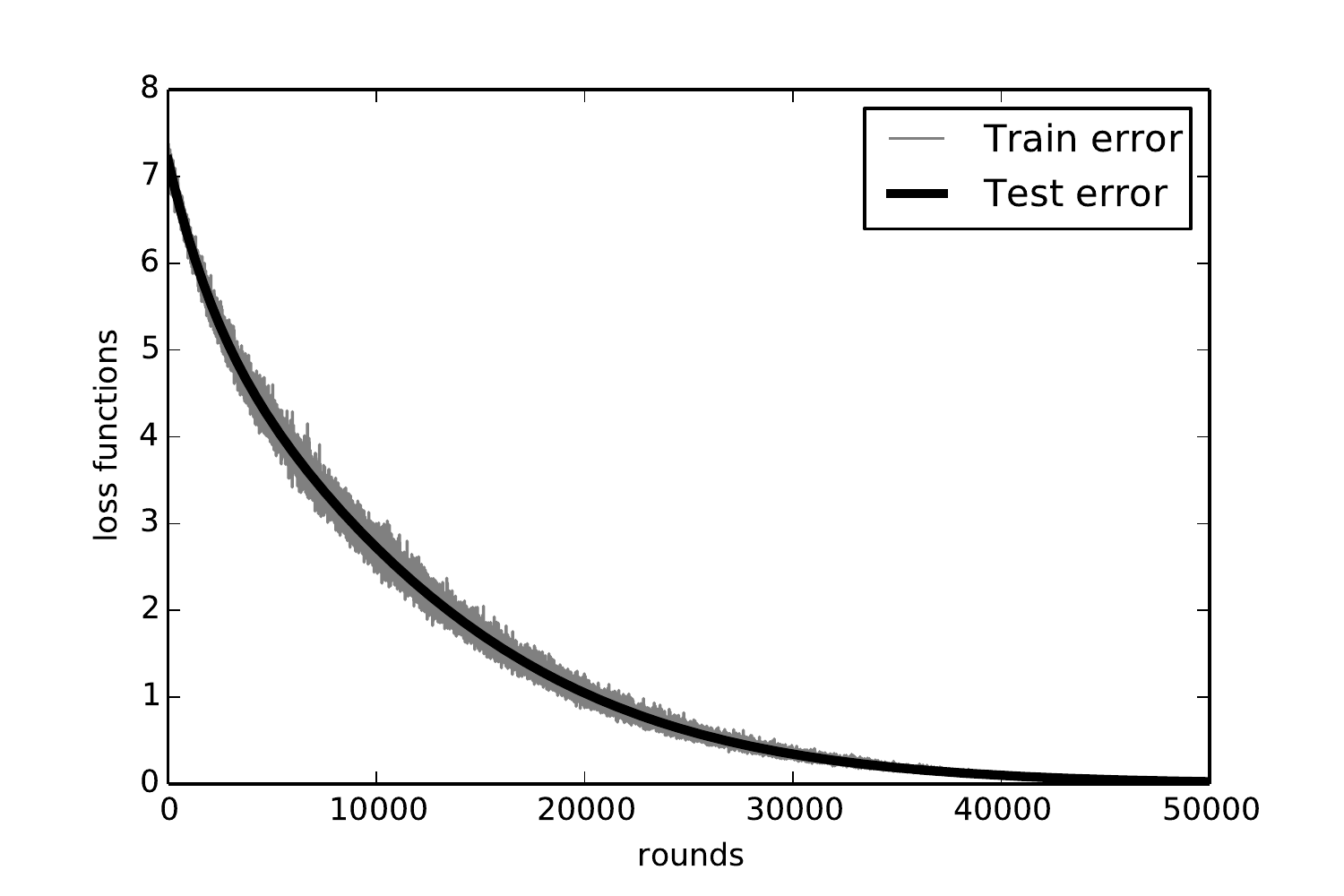}
                \caption{}
                \label{fig:test_errors}
        \end{subfigure}
        ~ 

                  \caption{(a) Denoising Auto-encoder (DAE) model. (b) Miss-classification rate for trained DAEs with different number of hidden nodes. (c) Train and test errors for a DAE trained with 100 hidden nodes; batch size of 64 samples; 50,000 rounds; a learning rate of 1e-5; and clipping thresholds of $\delta^{(t)}=1$.}\label{fig:general}
\end{figure}

DAE seek to reconstruct original basket $\mathbf{x}$ from corrupted input basket $\mathbf{\tilde{x}}$ by \textit{encoding} input $\mathbf{\tilde{x}}$ into $\mathbf{h}\in\mathbb{R}^N$ through mapping $\mathbf{h}=f(\mathbf{W}^i\mathbf{\tilde{x}}+\mathbf{b}^i)$, with $\mathbf{W}^i\in\mathbb{R}^{N\times p}$ and $\mathbf{b}^i\in\mathbb{R}^N$; and then \textit{decoding} it back into the input space as $\mathbf{y}\in\mathbb{R}^p$ through $\mathbf{y}=g(\mathbf{W}^o\mathbf{h}+\mathbf{b}^o)$, with $\mathbf{W}^o\in\mathbb{R}^{p\times N}$ and $\mathbf{b}^o\in\mathbb{R}^p$ (see Figure \ref{DAE}.) Vector $\mathbf{h}$ is usually referred to as the hidden layer of the auto-encoder. For convenience, we bundle the model parameters into $\mathbf{\theta}=\{\mathbf{W}^i, \mathbf{b}^i, \mathbf{W}^o, \mathbf{b}^o\}$. 

The DAE model is, thus, composed of the encode and decode operations, combined as
\begin{equation}
\mathbf{y}=g(f(\mathbf{\tilde{x}})).
\end{equation}
In this paper, the activation functions $f$ and $g$ are chosen to be the hyperbolic tangent and the sigmoid function, respectively, applied element-wise.

A key hyperparameter of the model is the dimension $N$ of the hidden layer. In traditional (non-denoising) auto-encoders (AE), if $N$ is equal or larger than the size of the input, the auto-encoder is able to trivially learn the identity function, providing no data compression. In these traditional AE's, to extract the salient features of interest from the input space, either $N$ must be smaller than the size of the input, or we must use regularization on the model (e.g., weight decay) or on the learning algorithm (e.g., early-stop rules.) Because of the random noise added in DAE, even hidden layers with a dimension $N$ larger than that the size of the input are able to extract relevant information from the input. This is because the DAE is not learning the input itself but, instead, it is learning the conditional density $\mathbb{P}(\mathbf{x}|\mathbf{\tilde{x}})$. Hyperparamenter $N$ can be chosen using cross-validation (see Figure \ref{fig:missclassification}.)

\subsection{Corruption Mechanism}

In DAE models there is an stochastic corruption mechanism, defined by conditional distribution $C(\mathbf{\tilde{x}}|\mathbf{x})$, generating corrupted baskets $\mathbf{\tilde{x}}$. To define $C(\mathbf{\tilde{x}}|\mathbf{x})$ let us first define the empirical na\"\i ve distribution $\mathbf{\pi}(\mathbf{x})=\prod_{i=1}^p\pi_i$, where $\pi_i$ is the probability that item $i$ is contained in a particular basket ($\pi_i$ is usually referred to as the support of item(set) $i$ in the affinity-analysis literature.) Marginal probabilities $\pi_i$ are estimated as the observed frequency of the presence of item $i$ in training set $\mathcal{T}$. To obtain \textit{meaningful} corrupted inputs, we seek to corrupt items proportionally to their occurrence in the training set. Specifically, we define conditional-marginal distribution $C(\tilde{x}_i|\mathbf{x})$ as
\begin{equation}
C(\tilde{x}_i|\mathbf{x}) = \left\{ \begin{array}{ll}
         0 & \mbox{if $x_i=0\mbox{ or } u_i\leq \pi_i$};\\
        1 & \mbox{else},\end{array} \right.
\end{equation}
where $u_i\sim \mbox{Uniform}(0,1)$. We further reject generated corrupted baskets $\mathbf{\tilde{x}}$ with elements $\tilde{x}_i=0$ for every $i$.


\subsection{Loss function}

Loss function $L$ quantifies the degree to which we are dissatisfied with a model output. While we are ultimately interested in maximizing the rate at which we correctly identify missing items, such a counting-like function is, nevertheless, not an appropriate loss function for learning algorithms due to its non-differentiability. As a surrogate function, we will use the cross-entropy between baskets $\mathbf{x}=[x_1,\dots,x_p]$ and $\mathbf{\bar{x}}=[\bar{x}_1,\dots,\bar{x}_p]$ given by
\begin{equation}\label{loss_function}
L(\mathbf{x},\mathbf{\bar{x}})=-\sum_{i=1}^p\left[x_i\log(\bar{x}_i)+(1-x_i)\log(1-\bar{x}_i)\right].
\end{equation}
Note that minimizing loss function $L$ is equivalent to maximizing the log-likelihood of Bernoulli random variables $x_i$'s, each with parameter $\bar{x}_i$, as
\begin{equation}\label{cross-entropy}
\log p(\mathbf{x}|\mathbf{\bar{x}})=\log\prod_{i=1}^p {\bar{x}}_i^{x_i}(1-\bar{x}_i)^{1-x_i}.
\end{equation}
This result shows that the DAE model is, in fact, trained to learn conditional density $\mathbb{P}(\mathbf{x}|\mathbf{\tilde{x}})$.

The network is trained by searching the parameter space for the values that minimize the loss function $L$, i.e., by solving optimization problem
\begin{equation}\label{optimization}
\min_{\mathbf{\theta}} \sum_{\mathbf{x}\in\mathcal{T},\mathbf{\tilde{x}}|\mathbf{x}\sim C}L(\mathbf{x},g(f(\mathbf{\mathbf{\tilde{x}}}))).
\end{equation}


\subsection{Training the model}

Optimization Problem (\ref{optimization}) is usually solved using the back-propagation (BP) algorithm, a gradient-descent method that takes advantage of patterns appearing in the composition of the chain rule of the gradient of the connections between layers. With large data sets and high-dimensional parameter space, BP is usually performed using the stochastic gradient descent (SGD) method, in which at each learning step only a randomly-chosen subset of the original training set is utilized. Specifically, in this work we used the Adam algorithm \cite{kingma_adam:_2014}, an implementation of SGD with first- and second-order parameter's momentum on the learning rule (for further details the reader is referred to \cite{bengio_practical_2012}.)




During early stages of the training process, we may need to avoid \textit{exploiting} gradients, which arbitrarily enlarge the learning steps, jeopardizing convergence. To avoid this situation (and based on the discussion in \cite{pascanu_difficulty_2013}) we clip the gradient as follows
\begin{equation}
\nabla f\left(\mathbf{x}^{(t)}\right) = \left\{ \begin{array}{ll}
         \nabla f\left(\mathbf{x}^{(t)}\right) & \mbox{if $\left\Vert\nabla f\left(\mathbf{x}^{(t)}\right)\right\Vert\leq \delta^{(t)}$};\\
        \left(\delta^{(t)}/\left\Vert\nabla f\left(\mathbf{x}^{(t)}\right)\right\Vert\right)\cdot\nabla f\left(\mathbf{x}^{(t)}\right) & \mbox{else},\end{array} \right.
\end{equation}
where $\left\Vert\cdot\right\Vert$ is chosen to be the euclidean norm. We further require that $\delta^{(t)}\rightarrow 0$ as $t\rightarrow \infty$.

A seemingly more efficient alternative to direct training DAEs is to use \textit{walkback training}, a variant approach proposed in \cite{bengio_generalized_2013} in which the training set is augmented by including samples with corrupted basket $\mathbf{\tilde{x}}$ obtained based on the currently learned version of $\mathbb{P}_\mathbf{\theta}(\mathbf{x}|\mathbf{\tilde{x}})$; the additional samples are obtained from a generative mechanism as the one presented in Section \ref{generation}. The walkback training procedure is similar in spirit to Contrastive Divergence with $k$ MCMC steps, proposed for training RBMs in \cite{hinton_fast_2006}.


\subsection{Evaluating the model}

While the model is trained on minimizing loss function $L$, its performance is, however, evaluated using a miss-classification rate scheme on individual items. For this, the output $\mathbf{y}$ of the model is discretized into $\mathbf{\hat{y}}$ by the rule $\hat{y}_i=\mathbb{I}(y_i>\eta)$, for a  threshold $\eta\in[0,1]$, where $\mathbb{I}$ is the indicator function. Threshold $\eta$ is another hyperparameter of the model that can be tuned using cross-validation.




\section{Collaborative filtering}

Motivating the application of the DAE model described in Section~\ref{model} as a probabilistic CF is straightforward. In the retail context, original basket $\mathbf{x}$ corresponds to the set of items that the customer consciously and unconsciously wants, i.e.: items the customer consciously wants; the ones he forgot or did not know he wants; and the ones he picked but are somehow ``unusual'' among \textit{similar} customers. Corruption process $C(\mathbf{\tilde{x}}|\mathbf{x})$ models the limitations of our human capabilities to either remember or identify every item we do want. Given observed basket $\mathbf{\tilde{x}}$, the DAE can compute probabilities for each possible basket $\mathbf{x}$ based on learned conditional distribution $\mathbb{P}_\mathbf{\theta}(\mathbf{x}|\mathbf{\tilde{x}})$, thus providing a set of probabilistic recommendations. 

\begin{figure}[ht]
        \centering
        \begin{subfigure}[h]{.35\textwidth}
                \includegraphics[width=\textwidth]{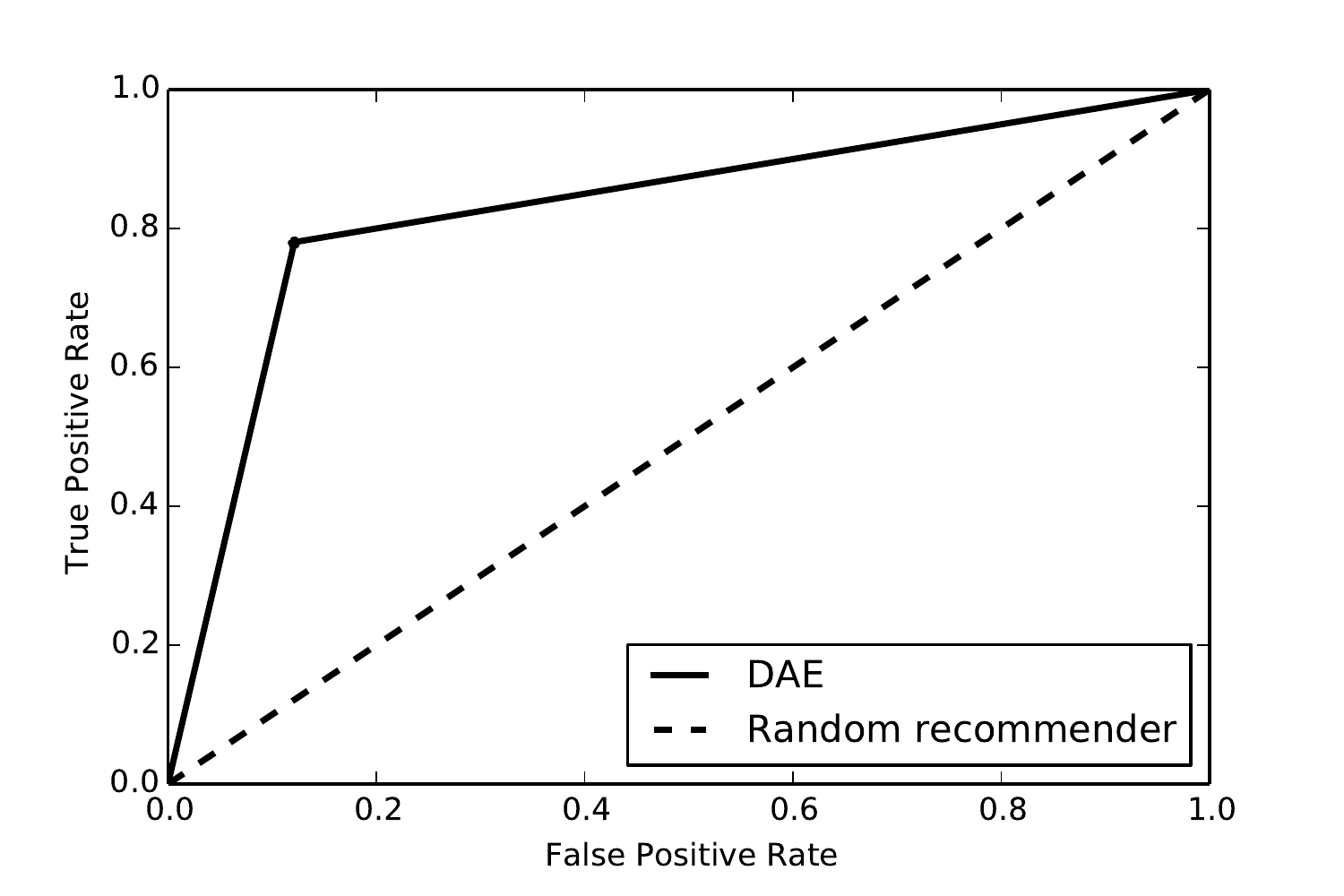}
                \caption{}
                \label{fig:ROC}
        \end{subfigure}
        ~ 
        ~ 
        \begin{subfigure}[h]{.35\textwidth}
                \includegraphics[width=\textwidth]{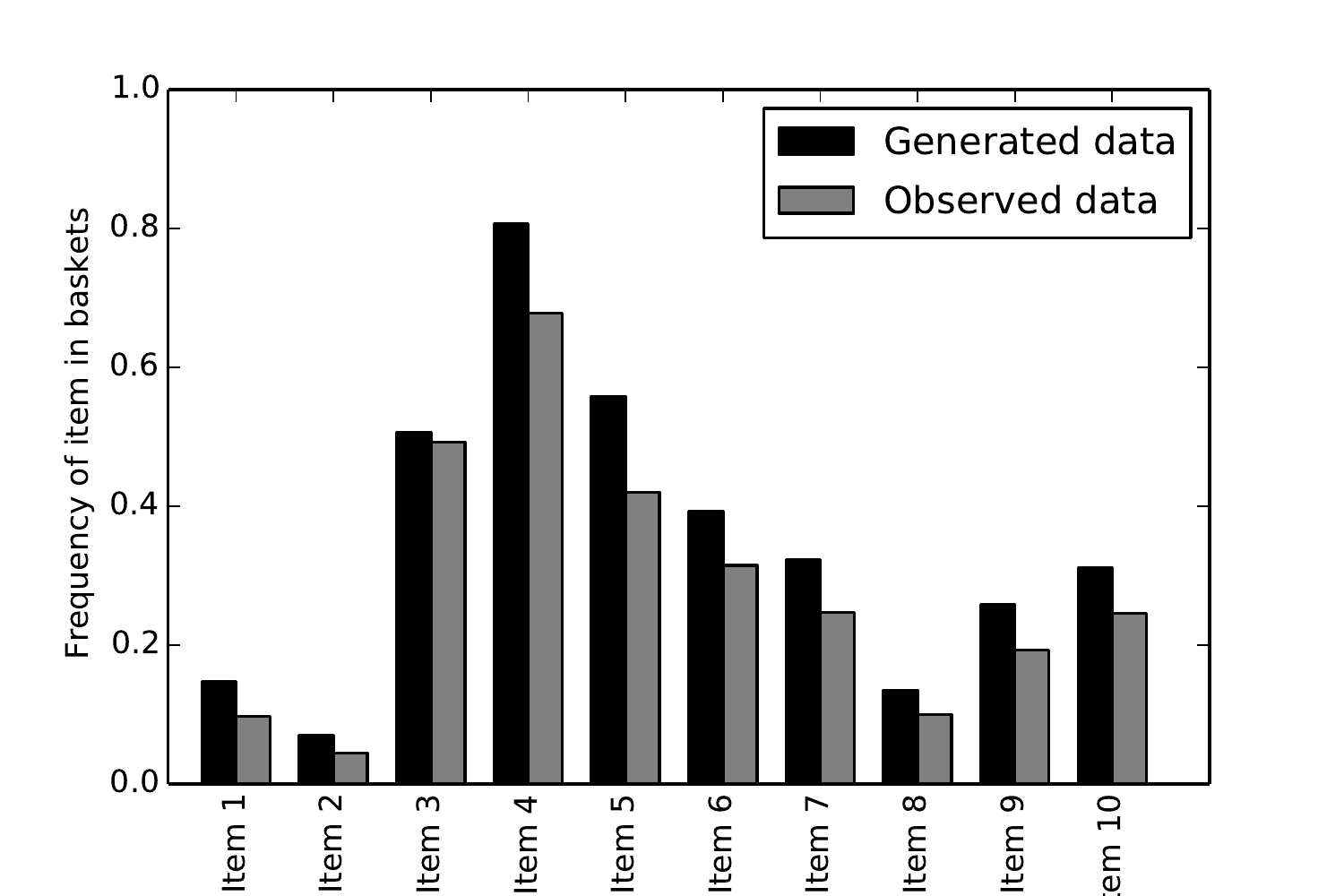}
                \caption{}
                \label{fig:frecuencies}
        \end{subfigure}
            \caption{(a) ROC curve for the DAE model. (b) Frequency of items in the training data set and from a generated data set obtained using Algorithm \ref{alg:main}.}\label{fig:general2}
\end{figure}

Using our trained DAE (see Figure \ref{fig:test_errors} and its caption) we evaluate its performance for recommendations using the ROC curve shown in Figure \ref{fig:ROC}. As it can be seen in the figure, points mainly concentrate around a False Positive Rate (FPR) of 0.12 and a True Positive Rate (TPR) of 0.77. Note that for a CF application we should mainly consider the FPR performance.

Table \ref{my-label} shows the confusion matrix for our DAE model. The matrix was obtained by considering the DAE-based CF performance on six thousand randomly-chosen baskets. The corresponding FPR is 0.122 and the TPR is 0.784.

\begin{table}[ht]
\centering
\caption{Confusion matrix for a DAE-based CF.}
\label{my-label}
\begin{tabular}{rccc}
\multicolumn{1}{c}{} & Predicted Missing & Predicted Present & Total  \\ \hline
Observed Missing     & 16,702            & 4,732             & 21,434 \\
Observed Present     & 4,601             & 33,965            & 38,566 \\ \hline
Total                & 21,303            & 38,697            & 60,000
\end{tabular}
\end{table}

\section{Market-Basket Generative Model}
\label{generation}


In this section we describe an iterative procedure to sample from distribution of baskets $\mathbb{P}(\mathbf{x})$ using a trained DAE. The procedure works by first sampling $\mathbf{\tilde{x}}^{(t)}$ from $C(\mathbf{\tilde{x}}|\mathbf{x}^{(t)})$; then sampling $\mathbf{x}^{(t+1)}$ from $\mathbb{P}_\mathbf{\theta}(\mathbf{x}|\mathbf{\tilde{x}}^{(t)})$; and repeat alternating. An initial point $\mathbf{x}^{(0)}$ may be picked arbitrarily. This procedure produces a homogeneous markov chain with transition probabilities $\mathbf{T}(\mathbf{x}^{(t+1)}|\mathbf{x}^{(t)})=\int \mathbb{P}_\mathbf{\theta}(\mathbf{x}^{(t+1)}|\mathbf{\tilde{x}})C(\mathbf{\tilde{x}}|\mathbf{x}^{(t)})d\mathbf{\tilde{x}}$. It was shown in \cite{bengio_generalized_2013} that if $\mathbb{P}_\mathbf{\theta}(\mathbf{x}|\mathbf{\tilde{x}})$ is a consistent estimator of $\mathbb{P}(\mathbf{x}|\mathbf{\tilde{x}})$, then the asymptotic marginal distribution of $\mathbf{T}$ (if it exists) is a consistent estimator of $\mathbb{P}(\mathbf{x})$. Algorithm \ref{alg:main} provides a methodology for generating random baskets with a convergent distribution $\mathbb{P}(\mathbf{x})$ using the afore-mentioned result.

\begin{algorithm}
\caption{Algorithm for Generating Random Baskets}
\label{alg:main}
\KwIn{Dataset $\mathcal{X}=\{\mathbf{x}_1,\dots,\mathbf{x}_N\}$ consisting of $N$ baskets}
\KwOut{Randomly-generated basket $\mathbf{x}$ from distribution $\mathbb{P}(\mathbf{x})$}
Sample $\mathbf{x}^{(t+1)}\sim\mathbf{\pi}(\mathbf{x})$\\
\Repeat{convergence}{
$\mathbf{x}^{(t)}=\mathbf{x}^{(t+1)}$\\
Sample $\mathbf{\tilde{x}}^{(t)}|\mathbf{x}^{(t)}\sim C(\mathbf{\tilde{x}}|\mathbf{x}^{(t)})$\\
Sample $\mathbf{x}^{(t+1)}|\mathbf{\tilde{x}}^{(t)}\sim \mathbb{P}_\mathbf{\theta}(\mathbf{x}|\mathbf{\tilde{x}}^{(t)})$\\
}
\Return Basket $\mathbf{x}^{(t+1)}\sim\mathbb{P}(\mathbf{x})$
\label{system}
\end{algorithm}

Sampling $\mathbf{x}^{(t+1)}\sim\mathbf{\pi}(\mathbf{x})$ (line 1 of Algorithm \ref{alg:main}) can be readily done by randomly sampling a basket from data set $\mathcal{S}$. On the other hand, sampling $\mathbf{x}^{(t+1)}|\mathbf{\tilde{x}}^{(t)}\sim \mathbb{P}_\mathbf{\theta}(\mathbf{x}|\mathbf{\tilde{x}}^{(t)})$ (line 5 of Algorithm \ref{alg:main}) can be achieved by sampling $\tilde{x}^{(t+1)}_i$ from a Bernoulli distribution with parameter $y_i^{(t)}=[g(f(\mathbf{x}^{(t)}))]_i$, for $i=1,\dots,p$. Figure \ref{fig:frecuencies} shows the item's frequency from the original data set and from a generated data set obtained using Algorithm \ref{alg:main}.

While the procedure described here resembles that of a Gibbs-sampling chain, it should be noted that it is not a proper one since there is no guarantee that $C(\mathbf{\tilde{x}}|\mathbf{x})$ and $\mathbb{P}_\mathbf{\theta}(\mathbf{x}|\mathbf{\tilde{x}})$ coherently describe a unique joint distribution. The procedure is, in fact, similar to the one proposed in \cite{heckerman_dependency_2000} for training dependency networks.

\section{Conclusions}

We presented a CF model obtained by training a DAE seeking to recover corrupted baskets. We showed that after careful tunning of the parameters, good results can be obtained and the DAE model used as a recommendation mechanism. Based on the DAE trained, a generative model was also presented, which can be used to generate synthetic baskets. This may be used for direct simulation analysis or as part of a more complex analytical method.

\section*{Acknowledgements}
The authors are thankful to the Board of Directors of Tiendas Industriales Asociadas Sociedad An\'onima (TIA S.A.), a leading supermarket retail chain in Ecuador, for authorizing their company to provide historical sales data for hundreds of their products and allowing invaluable discussions with their personnel, which help to shape the present research effort.

%

\bibliography{ID3251}

\end{document}